\documentclass[conference]{ieeeconf}
\usepackage{times}
\usepackage{amsmath}
\usepackage[pdftex]{graphicx}
\usepackage{subfigure}
\usepackage{amsmath,amssymb,amsopn,amstext,amsfonts}
\usepackage{cancel}
\usepackage[space]{cite}
\usepackage{pdfsync}
\usepackage{balance}
\usepackage{color}
\usepackage{algorithm}
\usepackage{algorithmic}
\usepackage{url}
\usepackage{booktabs}
\usepackage{threeparttable}

\usepackage[linkcolor=black,citecolor=black,urlcolor=black,colorlinks=true]{hyperref}
\usepackage{multirow}
\usepackage{epstopdf}
\usepackage{graphicx}
\IEEEoverridecommandlockouts

\newcommand\inv[1]{#1\raisebox{1.15ex}{$\scriptscriptstyle-\!1$}}

\graphicspath{{figures/}}
\DeclareGraphicsExtensions{.pdf,.png,.jpg,.eps}

\title{\LARGE \bf Relocalization, Global Optimization and Map Merging \\ for Monocular Visual-Inertial SLAM}
\author{Tong Qin, Peiliang Li, and Shaojie Shen
\thanks{All authors are with the Department of Electronic and Computer Engineering,
        Hong Kong University of Science and Technology, Hong Kong, China.
        {\tt\small \{tong.qin, pliap\}@connect.ust.hk, eeshaojie@ust.hk}.
        This work was supported by the Hong Kong Research Grants Council Early Career Scheme under project no. 26201616, and HKUST Proof-of-Concept Fund under project no. PCF.009.16/17.}
}

\begin{document}

\maketitle
\thispagestyle{empty}
\pagestyle{empty}

\begin{abstract}
The monocular visual-inertial system (VINS), which consists one camera and one low-cost inertial measurement unit (IMU), is a popular approach to achieve accurate 6-DOF state estimation. 
However, such locally accurate visual-inertial odometry is prone to drift and cannot provide absolute pose estimation. 
Leveraging history information to relocalize and correct drift has become a hot topic.
In this paper, we propose a monocular visual-inertial SLAM system, which can relocalize camera and get the absolute pose in a previous-built map.
Then 4-DOF pose graph optimization is performed to correct drifts and achieve global consistent. 
The 4-DOF contains x, y, z, and yaw angle, which is the actual drifted direction in the visual-inertial system. 
Furthermore, the proposed system can reuse a map by saving and loading it in an efficient way.
Current map and previous map can be merged together by the global pose graph optimization. 
We validate the accuracy of our system on public datasets and compare against other state-of-the-art algorithms. 
We also evaluate the map merging ability of our system in the large-scale outdoor environment. 
The source code of map reuse is integrated into our public code, VINS-Mono\footnote{https://github.com/HKUST-Aerial-Robotics/VINS-Mono}.

\end{abstract}

\section{Introduction}
Accurate state estimation plays an important role in a wide range of applications, such as autonomous navigation, virtual reality (VR), and augmented reality (AR). 
The camera has become a more and more popular sensor in this area.
Approaches~\cite{ForPizSca1405, engel2014lsd, mur2015orb, engel2017direct} that use a single camera has attracted significant attention. 
However, the metric scale cannot be directly recovered from one camera, which limits their usage in real world.
Recently, assisting the monocular camera with a low-cost inertial measurement unit (IMU) has become a popular trend.
IMUs measure acceleration and angular velocity, which render the scale, roll, and pitch angles observable.
Furthermore, the integration of IMU measurements can dramatically improve visual tracking performance in the texture-less area and aggressive motion, which extends the range of applications.
The monocular camera and IMU form the minimum sensor set for accurate and robust 6-DOF state estimation.

Due to the computation limitation and real-time requirement, visual-inertial odometry approaches only focus on local accuracy.
They process measurements collected within a local area or a short period while throw or marginalize past measurements.
Therefore, these approaches are prone to drift in a long run. 
Even IMUs can correct drifts in roll and pitch direction, the visual-inertial system still drifts in other four directions, x, y, z and yaw angle.
However, a globally drift-free trajectory is required in many tasks, such as robot exploration and navigation, and indoor augmented reality applications.
Therefore, it is of crucial importance for a SLAM system to have the ability to relocalize and correct drift smoothly. 

Another issue is that the visual-inertial odometry is a relative transformation from the initial frame instead of an absolute position. 
Every time we launch the system, it sets the start point as the reference frame and outputs the odometry in unfixed reference frames. 
Therefore, we cannot get poses in a consistent global frame in different trials. 
However, in some stable environments, we want to get the absolute pose in a fixed frame whenever and wherever we launch the system. 

\begin{figure}
    \centering
    \includegraphics[width=0.48\textwidth]{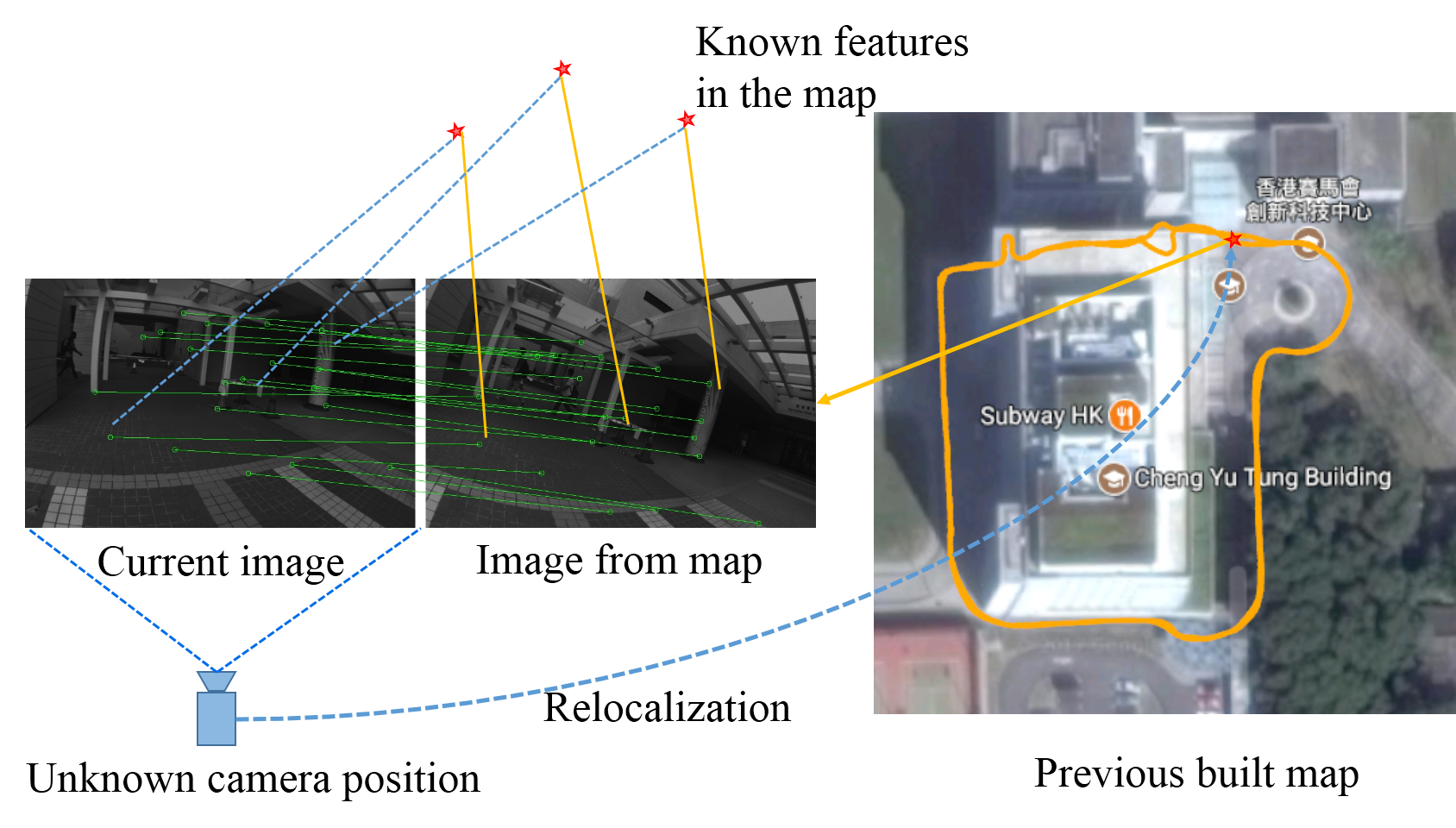}
    \caption{The proposed visual-inertial system relocalize camera position in real time. The right image is previous-built map aligned with Google Map. The camera starts at an unknown position. We retrieve descriptors on the current image. If a similar image view is found in the map, we relocate camera and get the absolute pose in the previous map. 
        \label{fig:abstract_image}}
\end{figure}

To address all these issues, we propose a real-time monocular visual-inertial system, which can achieve relocalization and global pose graph optimization to eliminate drift. 
Meanwhile, our system can reuse previous-built map and relocalize current pose in the previous-built map. 
Therefore, we can get the absolute pose estimation in a known environment. 
Furthermore, our system can merge current map into previous-built map smoothly. 
Our system is based on a real-time monocular visual-inertial odometry (VIO) method which provides locally accurate estimation \cite{QinShen17}. 
Loop detection is achieved by a state-of-the-art image retrieval method, DBoW2 \cite{GalvezTRO12}. 
Relocalization is done in a tightly-coupled feature-level fusion with the monocular VIO. 
Finally, geometrically verified loops are added into a 4-DOF pose graph optimization to eliminate drift smoothly. 
The experiments show that the proposed system can improve localization accuracy. 
Also, the map "evolves" overtime by incrementally merging new sensor data captured at different times.

We highlight that our contribution in twofold:
\begin{itemize}
\item A complete SLAM system with relocalization, 4-DOF pose graph optimization, map merging and reuse of previous-built map.
\item Open-source code of map reuse.  
\end{itemize}

The rest of the paper is structured as follows. 
In Sect.~\ref{sec:literature}, we discuss the relevant literature. 
The system overview is discussed in Sect.~\ref{sec:System Overview}. 
We introduce our algorithm in detail in Sect.~\ref{sec:algorithm}.
Implementation details and experimental evaluations are presented in Sect.~\ref{sec:experiments}. 
Finally, the paper is concluded in Sect.~\ref{sec:conclusion}.

\section{Related Work}
\label{sec:literature}
Tremendous research works on visual SLAM have appeared in the last few years.
Current state-of-the-art monocular approaches include  SVO~\cite{ForPizSca1405}, LSD-SLAM~\cite{engel2014lsd},DSO~\cite{engel2017direct}, ORB-SLAM~\cite{mur2015orb} and so on. 
They use monocular vision to track camera pose and map the environment at the same time.
Some of them are based on sparse features and some of them are based on the dense image.
They achieve convincing results of localization and mapping in an up-to-scale structure.

In order to recover the real scale, IMU is often used to assist camera in the visual system.
IMU-aided visual odometry has attracted significant attention recently. 
Some simple but effective works ~\cite{weiss2012real,lynen2013robust} loosely fuse IMU and camera by Kalman Filter (KF). 
The visual result is independent of IMU.  
The camera depicts the up-to-scale structure firstly, then the IMU complement the scale.
Tightly-coupled visual-inertial fusion can achieve higher accuracy.
One popular EKF based VIO approach is MSCKF~\cite{MouRou0704,LiMou1305}. 
Several camera poses are maintained in the state vector.
Therefore, the observations of the same features crossing multiple camera views form the multi-constraint update.
The camera poses, velocity and IMU bias are jointly updated.
SR-ISWF \cite{wu2015square,paulcomparative} is a similar work with MSCKF. 
The improvement is that it uses square-root form \cite{kaess2012isam2} to achieve single-precision representation and avoid poor numerical properties, which can run on computation-limited platforms, such as mobile devices. 
Another trend uses graph optimization ~\cite{LeuFurRab1306,SheMicKum1505,yang2017monocular,QinShen17} to tightly solve the visual-inertial problem. 
They usually keep multiple camera measurements and IMU measurements in a bundle and jointly optimize them to obtain the optimal state estimates.
The graph optimization framework usually requires high computation resource. 
To bound computation complexity and achieve real-time performance, 
some visual-inertial odometry methods~\cite{LeuFurRab1306,yang2017monocular, QinShen17} keep a limited window size of recent states, while marginalize out past states.
Therefore, these approaches focus on local accuracy. 
Accumulating drift is unavoidable in a long run.

Relocalization algorithms can be divided into two categories based on the type of map.
One is the offline-built map, and the other one is the online-built map. Noticeable works based on the offline-built map include \cite{kendall2015posenet, lynen2015get, qiu2017model}.  \cite{lynen2015get, qiu2017model} build an offline map in geometric configuration while \cite{kendall2015posenet} build the offline map by learning method.
\cite{burri2015real, mur2017visual, kasyanov2017keyframe} are algorithms which can achieve relocalization in the visual-inertial system with an online-built map. \cite{burri2015real} and \cite{kasyanov2017keyframe} retrieve previous images by BRISK \cite{leutenegger2011brisk} features. 
Raul et al. \cite{mur2017visual} retrieves previous images by ORB \cite{rublee2011orb} features.
Burri et al. \cite{burri2015real} and Raul et al \cite{mur2017visual} achieve global consistency by a global bundle adjustment (BA) in the background, while Kasyanov et al. \cite{kasyanov2017keyframe} achieves this by 6-DOF pose graph optimization. 
However, these online map building algorithms lack the ability to load and reuse a previous build map.
Also, in contrast with Kasyanov et al. \cite{kasyanov2017keyframe}, we perform 4-DOF pose graph optimization on 3D translation and the rotation around the gravity direction (yaw angle), which are minimum unobservable directions.

\section{System Overview}
\label{sec:System Overview}

\begin{figure}
    \centering
    \includegraphics[width=0.48\textwidth]{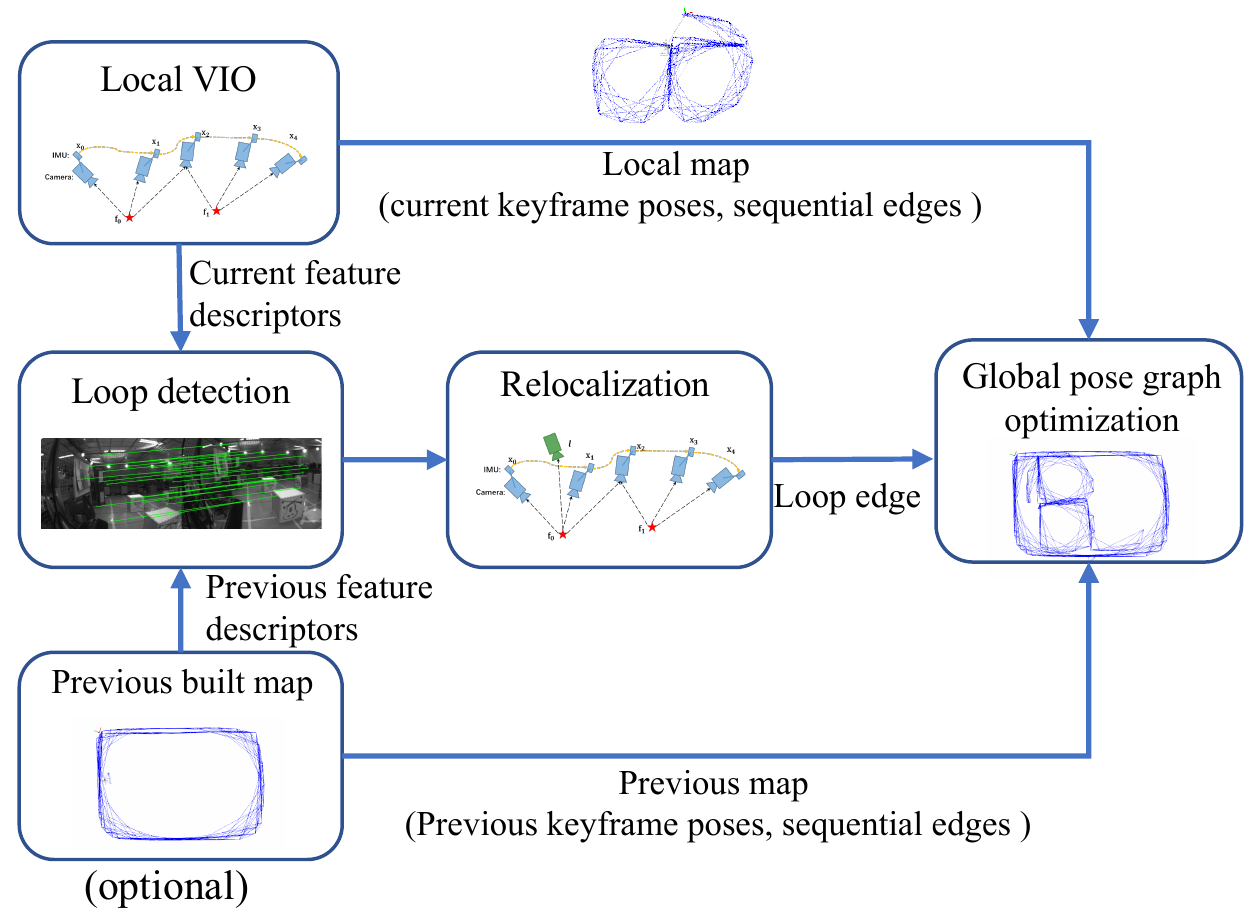}
    \caption{A block diagram illustrating the full pipeline of the proposed monocular visual-inertial system.
        \label{fig:system}}
     \vspace{-0.5cm}
\end{figure}

The pipeline of the proposed visual-inertial system is depicted in Fig. \ref{fig:system}. 
The system starts with a state-of-the-art monocular visual-inertial odometry \cite{QinShen17}, which achieves high accuracy local estimation. The VIO method keeps several keyframes in the local window and marginalized past frames. 
The keyframe poses are added into a global pose graph, which runs in another thread.
Meanwhile, we process relocalization procedure on every keyframe in the third thread. 
The relocalization process starts with a loop detection module that identifies places that have already been visited.
Once current keyframe detects loop with the previous keyframe, relocalization is processed immediately by jointly optimizing previous keyframe with the local window in raw feature-level. 
Then the loop information will be added to pose graph as a loop edge connecting current keyframe with the loop closure frame. 
If we have a map that is previously built, we can directly load it into the pose graph. 
The global pose graph thoroughly optimizes all edges from current and previous-built map to achieve global consistency. 
In the consideration of large-scale environment, we only maintain a sparse pose graph in our map, instead of the full bundle.

\begin{figure}[h]
    \centering
    \includegraphics[width=0.4\textwidth]{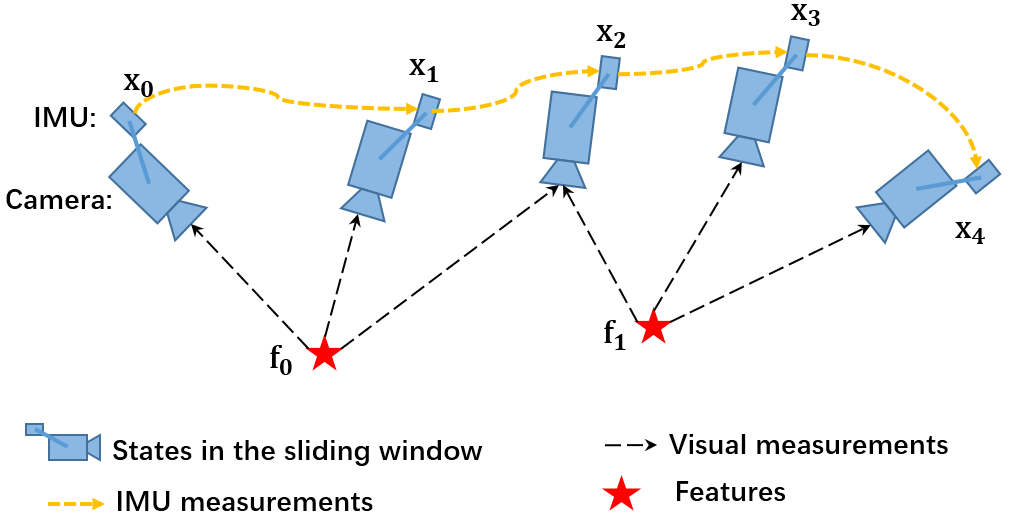}
    \caption{An illustration of sliding-window based monocular VIO. The local window keeps several keyframes and IMU measurements between consecutive keyframes. A local bundle adjustment (BA) jointly optimization keyframes poses, velocity, IMU bias as well as feature depths. 
        \label{fig:vio_f} }
    \vspace{-0.5cm}
\end{figure}

\section{Algorithm}
\label{sec:algorithm}

\subsection{Visual-inertial Odometry}
\label{sec:Visual-inertial Odometry}

We adopt the algorithm proposed in \cite{QinShen17} for monocular visual-inertial odometry. 
As depicted in the Fig. \ref{fig:vio_f}, The sliding window based nonlinear optimization framework processes visual and inertial measurements in a tightly-coupled way. 
Conner features are detected \cite{ShiTom9406} and tracked \cite{LucKan8108}, while IMU measurements are locally integrated.
The VIO starts with a robust initialization procedure to guarantee the system can launch under any unknown state or motion. 
Poses, velocities, IMU bias of several keyframes as well as feature depths are optimized in a local bundle adjustment. 
Only keyframes, which contain sufficient feature parallax with its neighbors, are temporarily kept in the local window. 
Previous keyframes are marginalized out of the window in order to bound computation complexity.

The definition of full states in a sliding window with $n$ frames and $m$ features are (the transpose is ignored):
\begin{equation}
\begin{split}
\mathcal{X}    &= \left [ \mathbf{x}_0,\,\mathbf{x}_{1},\, \cdots \,\mathbf{x}_{n},\, \mathbf{x}^b_c,\, \lambda_0,\,\lambda_{1},\, \cdots \,\lambda_{m} \right ] \\
\mathbf{x}_k   &= \left [ \mathbf{p}^w_{b_k},\,\mathbf{v}^w_{b_k},\,\mathbf{q}^w_{b_k}, \,\mathbf{b}_a, \,\mathbf{b}_g \right ], k\in [0,n] \\
\mathbf{x}^b_c  &= \left [ \mathbf{p}^b_c,\,\mathbf{q}^b_{c} \right ],
\end{split}
\end{equation}
where the $k$-th IMU state consists of the position $\mathbf{p}^{w}_{b_k}$, velocity $\mathbf{v}^{w}_{b_k}$, orientation $\mathbf{q}^{w}_{b_k}$ in the world frame, and IMU bias $\mathbf{b}_a$, $\mathbf{b}_g$ in body frame. 3D features are parameterized by their inverse depth $\lambda$ when first observed in camera frame, and $\mathbf{x}^b_c$ is the extrinsic transformation from camera frame $c$ to body frame $b$.
The estimation is formulated as a nonlinear least-square problem:
\begin{equation}
\label{eq:nonlinear_cost_function}
\begin{aligned}
\min_{\mathcal{X}} \left \{ \left 
\| \mathbf{r}_p - \mathbf{H}_p \mathcal{X} \right \|^2 
+ \sum_{k \in \mathcal{B}}  \left \| \mathbf{r}_{\mathcal{B}}(\hat{\mathbf{z}}^{b_k}_{b_{k+1}},\, \mathcal{X}) \right \|_{\mathbf{P}^{b_k}_{b_{k+1}}}^2 + \right. \\
\left. 
\sum_{(l,j) \in \mathcal{C}} \rho( \left \| \mathbf{r}_{\mathcal{C}}(\hat{\mathbf{z}}^{c_j}_l ,\, \mathcal{X}) \right \|_{\mathbf{P}^{c_j}_l}^2 )
\right\},
\end{aligned}                    
\end{equation}
where $r_{\mathcal{B}}(\hat{\mathbf{z}}^{b_k}_{b{k+1}},\, \mathcal{X})$ and $r_{\mathcal{C}}(\hat{\mathbf{z}}^{c_j}_l,\, \mathcal{X})$ are nonlinear residual functions for inertial and visual measurements. 
$||\cdot ||$ is the Mahalanobis distance weighted by covariance $\mathbf{P}$. 
To be specific, $r_{\mathcal{B}}$ is the residual of IMU factor which constrains pair of consecutive frames $b_{k}$ and $b_{k+1}$ by the integration of inertial measurements $\hat{\mathbf{z}}^{b_k}_{b_{k+1}}$. 
$r_{\mathcal{C}}$ is the residual of vision factor which represents the reprojection error by reprojecting feature $l$ into frame $j$ and comparing against raw measurements $\hat{\mathbf{z}}^{c_j}_l$. 
$\rho (\cdot)$ is the robust huber norm \cite{Hub64} to relieve outliers.
Past states are marginalized and converted to the prior information, $\{\mathbf{r}_p,\,\mathbf{H}_p\}$.

Only a small set of recent frames is optimized in the window, and the past states are linearized and fixed into marginalization factor. 
Therefore, accumulating drift is inevitable in a long run.

\begin{figure}
    \centering
    \subfigure[BRIEF descriptor matching results]{
        \label{fig:matching0}    
        \includegraphics[width=0.9\columnwidth]{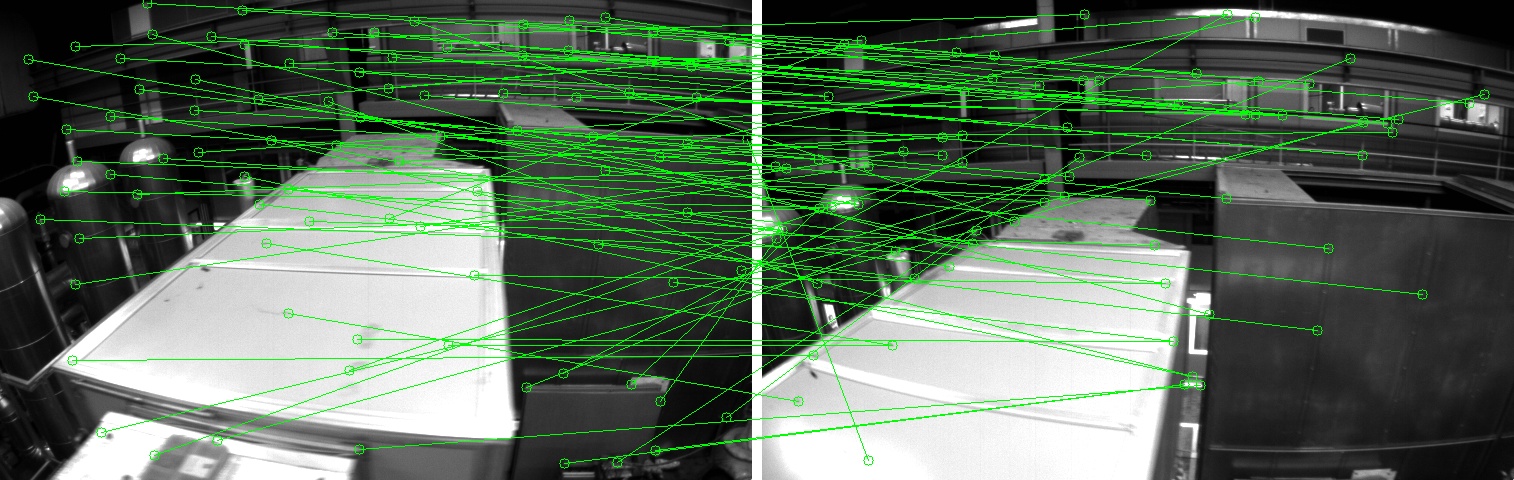}}   
    \subfigure[First step: 2D-2D outlier rejection results]{
        \label{fig:matching1}
        \includegraphics[width=0.9\columnwidth]{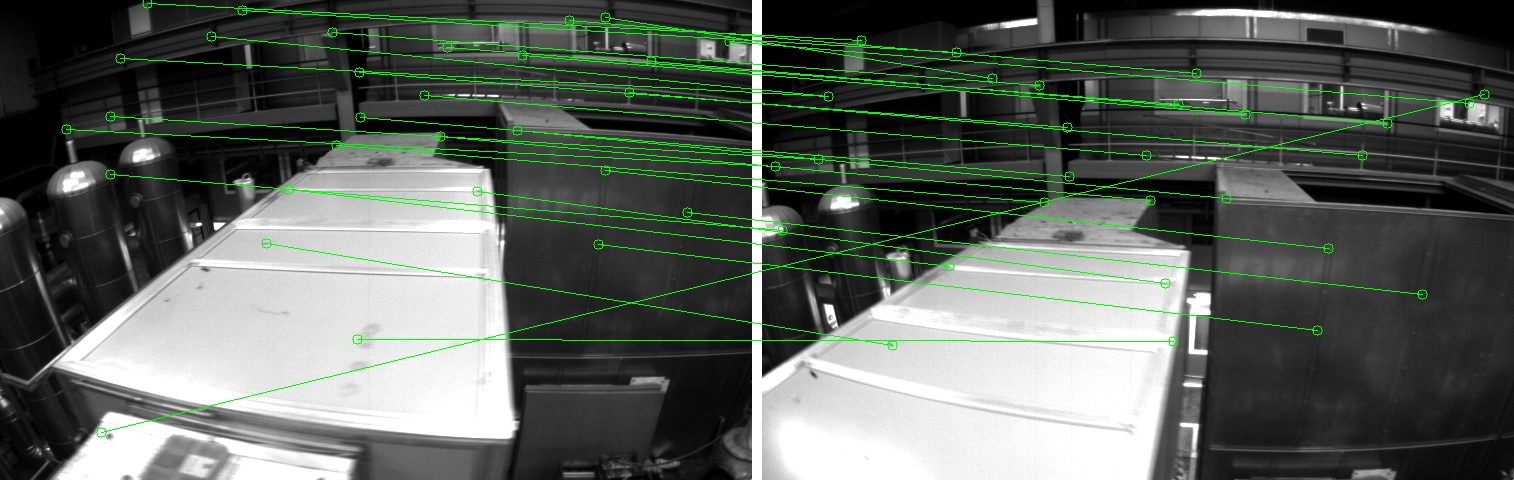}}
    \subfigure[Second step: 3D-2D outlier rejection results.]{
        \label{fig:matching2}
        \includegraphics[width=0.9\columnwidth]{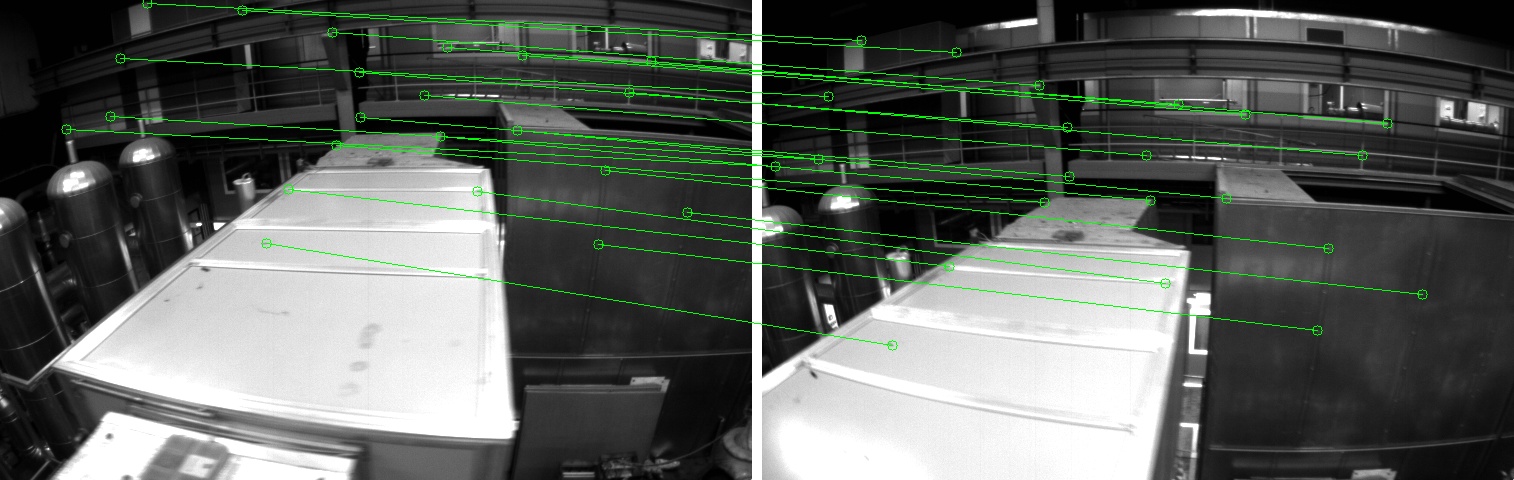}}
    \caption{Descriptor matching and outlier removal for feature retrieval during loop detection.}
    \label{fig:matching}
    \vspace{-0.5cm}
\end{figure}

\subsection{Loop Detection}
\label{sec:Loop Detection}
To achieve relocalization, we need to identify places that have already been visited.
We follow a state-of-the-art approach DBOW2~\cite{GalvezTRO12} for loop detection.
For every keyframe, we detect 500 FAST features~\cite{rosten2006machine} and describe them by the BRIEF descriptors~\cite{calonder2010brief}.
The descriptors are converted to a visual vector to query the visual database. 
We get the best loop closure candidate from DBOW2. 
The descriptors are also used for feature retrieving. 
Raw images are discarded to reduce memory.

\subsubsection{Feature Retrieval}
After loop detection, we establish the connection between the current frame and loop closure frame in feature level. 
The feature matching is performed by the BRIEF descriptors matching.
We choose the matching pairs by finding the minimum Hamming distance. 
Directly descriptor matching may cause a lot of outliers, as shown in Fig.~\ref{fig:matching}.
To remove outliers and verify the loop detection, we perform two-step geometric outlier rejection procedure.  
\begin{itemize}
    \item 2D-2D: We perform fundamental matrix test with RANSAC~\cite{hartley2003multiple} on 2D observation of matched pairs.
    
    \item 3D-2D: We perform PnP test with RANSAC~\cite{lepetit2009epnp} between 3D positions of features (from VIO) and 2D observations on the loop closure frame.
\end{itemize}
When we find enough inliers, we treat this candidate as a correct loop detection.

\subsection{Tightly-Coupled Relocalization}
\label{sec:relocalization}
\begin{figure}
    \centering
    \includegraphics[width=0.45\textwidth]{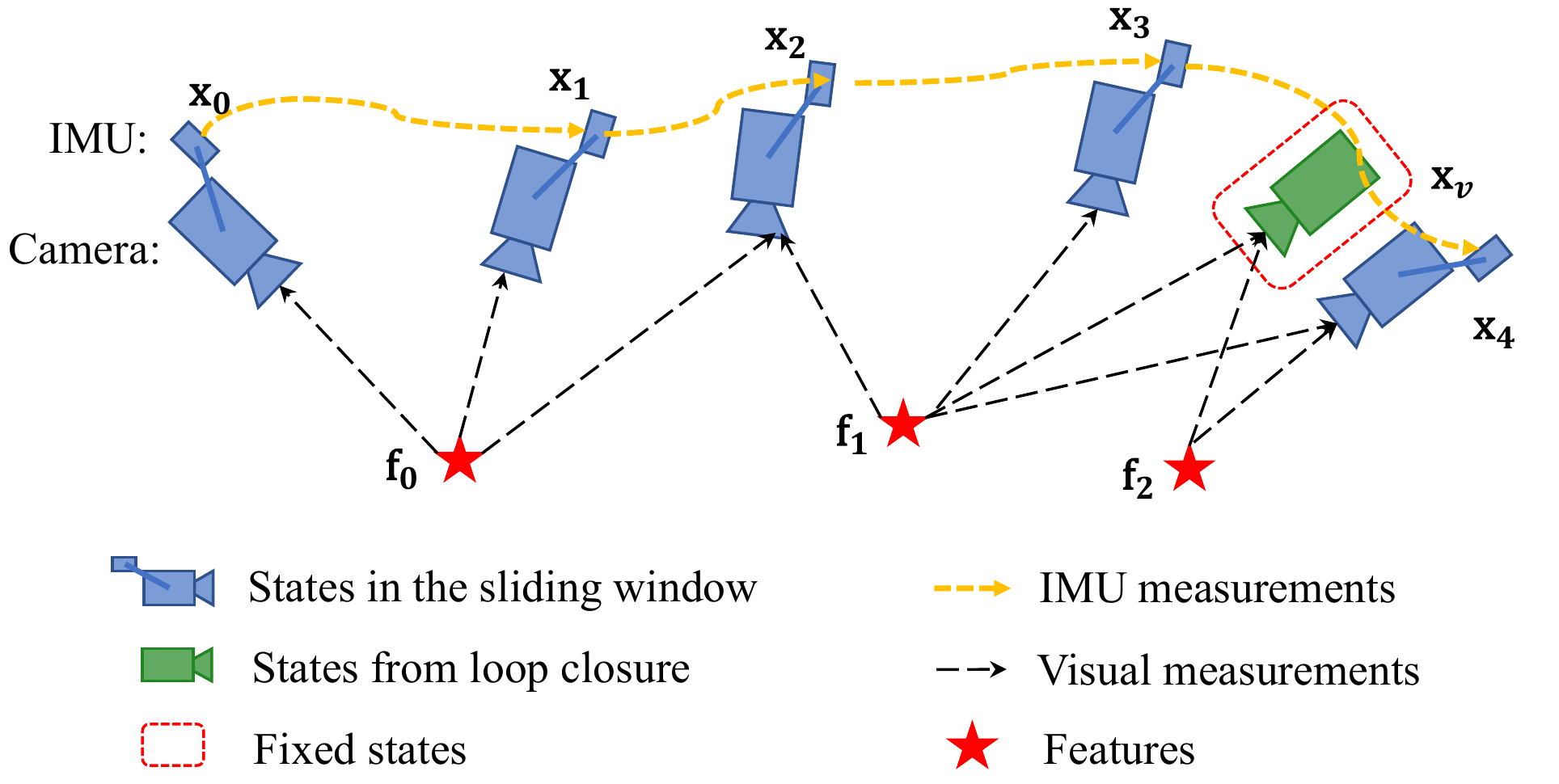}
    \caption{An illustration of sliding-window based monocular VIO with a loop closure frame. The loop closure frame serves as an additional camera view with the fixed pose in the local window. A local bundle adjustment (BA) jointly optimization keyframes poses, velocity, IMU bias as well as feature depths. 
        \label{fig:relocalize_m} }
    \vspace{-0.7cm}
\end{figure}
Instead of calculating relative pose just between two matched frames, we solve it by jointly optimize loop closure frame within the local window. 
The loop closure frame is treated as an additional frame with the fixed pose in the local sliding window of VIO, as shown in the Fig. \ref{fig:relocalize_m}. 
The connection is established by retrieved features.

We use $v$ to denote loop closure frame. 
During relocalization, we treat previous pose estimation ($\hat{\mathbf{q}}^w_v, \hat{\mathbf{p}}^w_v$) of frame $v$ as constant. 
The loop closure frame connects local window by retrieved features.
We jointly optimize the sliding window using all IMU measurements, local visual measurement measurements, and retrieved feature correspondences from loop closure.
We can easily write the same visual measurement model for retrieved features observed by a loop closure frame $v$. 
The nonlinear cost function in \eqref{eq:nonlinear_cost_function} only need to add visual reprojection error term of loop closure frame:
\begin{equation}
\begin{aligned}
\label{eq:jointly optimize loop}
\min_{\mathcal{X}}  \left\{ \left\| \mathbf{r}_p - \mathbf{H}_p \mathcal{X} \right\|^2
+ \sum_{k \in \mathcal{B}}  \left\| \mathbf{r}_{\mathcal{B}}(\hat{\mathbf{z}}^{b_k}_{b_{k+1}}, \mathcal{X})  \right\|_{\mathbf{P}^{b_k}_{b_{k+1}}}^2 \right. \\
\left. 
+ \sum_{(l,j) \in \mathcal{C}} \rho ( \left\| \mathbf{r}_{\mathcal{C}}(\hat{\mathbf{z}}^{c_j}_l, \mathcal{X})  \right\|_{\mathbf{P}^{c_j}_l}^2 ) \right. \\
\left.  
+ \underbrace{ \sum_{(l,v) \in \mathcal{L}} \rho ( \left\| \mathbf{r}_{\mathcal{C}}(\hat{\mathbf{z}}^{v}_l, \mathcal{X}, \hat{\mathbf{q}}^w_v, \hat{\mathbf{p}}^w_v) \right\|_{\mathbf{P}^{c_v}_l}^2 ) }_{\text{reprojection error in loop closure frame}}
\right\},
\end{aligned}                    
\end{equation}
where $\mathcal{L}$ is the set of the observation of retrieved features. 
${(l,v)}$ means $l^{th}$ feature observed in the loop closure frame $v$. 

This joint optimization framework results in higher accuracy.
The relocalization process effectively shifts local window to "drift-free" location. 

\subsection{Global Pose Graph Optimization}
\label{sec:pose graph optimization}

Since we fix past states in the relocalization procedure, the local window shifts to a "drift-free" place immediately. 
A jumping change will appear on the trajectory.
To make the whole trajectory consistent and smooth, we perform a light-weight 4-DOF global pose graph optimization. 

\subsubsection{Four Accumulated Drift Direction}
\begin{figure}
    \centering
    \includegraphics[width=0.4\textwidth]{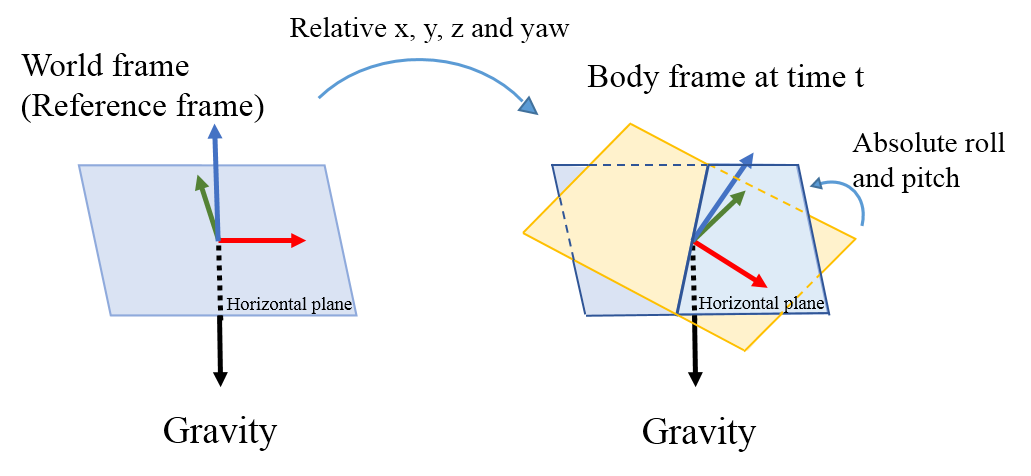}
    \caption{An illustration of four drifted direction. The measurement of gravity renders roll and pitch angle fully observable.  
        With the movement of the object, the x, y, z and yaw angle change relatively with respect to the reference frame. 
        The absolute roll and pitch angle can be determined by the horizontal plane from the gravity vector.
        \label{fig:4dof} }
     \vspace{-0.65cm}
\end{figure}

Since IMU measures gravity vector, the roll and pitch angles are fully observable in the visual-inertial system. 
As depicted in the Fig. \ref{fig:4dof}, the gravity is always in the vertical direction. 
With the movement of the object, the 3D position and rotation change relatively with respect to the reference frame. 
However, we can determinate horizontal plane by the gravity vectors, that means we can observable the absolute roll and pitch angles all the time. 
Therefore, the roll and pitch are absolute values in the world frame, while the x, y, z and yaw are relative estimates with respect to the reference frame. 
The accumulated drift only occurs in four degrees-of-freedom (x, y, z and yaw angle). 
To take full advantage of valid information and correct drift efficiently, we fix the drift-free roll and pitch, and only perform pose graph optimization in 4-DOF.

\subsubsection{Adding Keyframes into the Pose Graph}
\label{sec:Adding Keyframe into Pose Graph}
\begin{figure}
    \centering
    \includegraphics[width=0.45\textwidth]{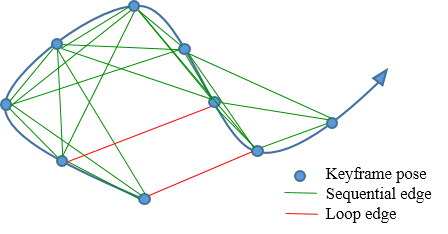}
    \caption{An illustration of pose graph. The keyframe serves as a vertex in the pose graph and it connects other vertexes by sequential edges and loop edges. Every edge represents relative translation and relative yaw angle.
        \label{fig:pose_graph} }
     \vspace{-0.2cm}
\end{figure}

Every keyframe is added into the pose graph after it is marginalized out from VIO local window.
One keyframe serves as one vertex in the pose graph. 
Every vertex connects others by two types of edges, sequential edge and loop edge, as shown in Fig. \ref{fig:pose_graph}:
\begin{itemize}
    \item Sequential Edge: 
    a keyframe will connect several previous keyframes with sequential edges. 
    The sequential edge represents the relative transformation between two vertexes, which is taken directly from VIO result. 
    Considering a keyframe $i$ and one of its previous keyframes $j$, the sequential edge contains relative position $\hat{\mathbf{p}}^i_{ij}$ in local frame and relative yaw angle $\hat{\psi}_{ij}$,
    \begin{equation}
    \label{eq:loop constraint}
    \begin{aligned}
    \hat{\mathbf{p}}^i_{ij} =& \inv{\hat{\mathbf{R}}^w_i}(\hat{\mathbf{p}}^w_j - \hat{\mathbf{p}}^w_i)\\
    \hat{\psi}_{ij} =& \hat{{\psi}}_{j} -  \hat{\psi}_{i}.
    \end{aligned}                    
    \end{equation}

    \item Loop Edge: 
    if loop detection happens, the keyframe will connect the loop closure frame by a loop edge. 
    Similar with the sequential edge, the loop edge also contains 4-DOF relative transformation.
    The value of the loop closure edge is obtained from relocalization result.
\end{itemize}

\subsubsection{4-DOF Pose Graph Optimization}
\label{sec:pose_graph}
The main idea of the pose graph optimization is that we adjust the poses of vertexes, such that the configuration matches the edges as much as possible. 
In our framework, we only adjust 3D position $\mathbf{p}^w$ and yaw angle $\psi$ of vertexes, and set their roll and pitch angles as constant variables. 
Such that loop correction will not occur in drift-free direction.
We define the residual of the edge between frame $i$ and $j$ as:  
\begin{equation}
\label{eq:pose graph residual}
\begin{split}
\mathbf{r}_{i,j}(\mathbf{p}^w_i, \psi_i, \mathbf{p}^w_j, \psi_j)=
\begin{bmatrix}
\inv{\mathbf{R}(\hat{\phi}_i,\hat{\theta}_i,\psi_i)} (\mathbf{p}^w_j - \mathbf{p}^w_i)-\hat{\mathbf{p}}^i_{ij}\\
\psi_j - \psi_i - \hat{\psi}_{ij}\\
\end{bmatrix}
\end{split},
\end{equation} 
where $\mathbf{p}^w_i, \psi_i, \mathbf{p}^w_j, \psi_j$ are variables of $i$ and $j$ frame. $\hat{\phi}_i, \hat{\theta}_i$ are the fixed roll and pitch angles obtained from VIO. $\hat{\mathbf{p}}^i_{ij},\hat{\psi}_{ij}$ are relative transformation from edge.

The residual of all sequential edges and loop closure edges are formed into following least squares problem:
\begin{equation}
\label{eq:pose graph optimization}
\begin{aligned}
\min_{\mathbf{p}, \psi} \left \{ 
\sum_{(i,j) \in \mathcal{S}} \left \| \mathbf{r}_{i,j} \right \|^2  + 
\sum_{(i,j) \in \mathcal{L}}   \rho (\left  \| \mathbf{r}_{i,j} \right \|^2 ) \right\},
\end{aligned}                    
\end{equation}
where $\mathcal{S}$ is the set of all sequential edges and $\mathcal{L}$ is the set of all loop closure edges. 
Huber norm $\rho(\cdot)$ is used to further reduce the impact of any possible wrong loops.

\subsubsection{Map Merging}
\begin{figure}
    \centering
    \includegraphics[width=0.45\textwidth]{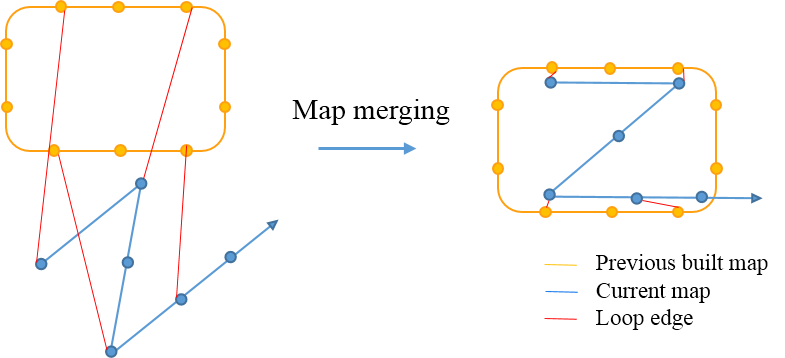}
    \caption{An illustration of map merging. The yellow figure is previous-built map. The blue figure is the current map. Two maps are merged according to the loop connections. 
        \label{fig:map_merge} }
     \vspace{-0.5cm}
\end{figure}

The pose graph can not only optimize current map, but also merge current map with a previous-built map.
If we have loaded a previous-built map and detected loop connections between two map, we can merge them together. 
Since all edges are relative constraints, the pose graph optimization automatically merges two maps together by the loop connections.
As shown in the Fig. \ref{fig:map_merge}, the current map is pulled into the previous map by loop edges. 
Every vertex and every edge are relative variables, therefore, we only need to fix the first vertex in the pose graph. 

\subsubsection{Pose Graph Management}
When the travel distance increases, the size of the pose graph may grow unbounded, limiting the real-time performance of the system.
To this end, we implement a downsample process to maintain the pose graph database to a limited size.
All keyframes with loop closure constraints will be kept, while other keyframes that are either too close or have very similar orientations to its neighbors are removed.

\subsection{Map Reuse}
In a stable environment, we can first build and save the map. 
Then we load and reuse the map for the next time. 
Current pose is relocated in the previous map and the current map is merged into the previous map.
Not only we achieve map reuse, but also we can always get the absolute odometry in this known environment.

\subsubsection{Pose Graph Saving}
The structure of our pose graph is very simple. We only need to save vertexes and edges, as well as descriptors of every keyframe (vertex). Raw images are discarded to reduce memory consumption. 
To be specific, the states we save for $i^{th}$ keyframe are:

\begin{equation}
[i ,\hat{\mathbf{p}}^w_i,\hat{\mathbf{q}}^w_i, v, \hat{\mathbf{p}}^i_{iv}, \hat{\psi}_{iv}, \mathbf{D}(u,v,des)], 
\end{equation}
where $i$ is frame index, $\hat{\mathbf{p}}^w_i$ and $\hat{\mathbf{q}}^w_i$ are position and orientation from VIO.
If this frame has a loop closure frame, $v$ is the loop closure frame's index.
$\hat{\mathbf{p}}^i_{iv}$ and $\hat{\psi}_{iv}$ are the relative position and yaw angle between these two frames, which is obtained from relocalization. 
$\mathbf{D}(u,v,des)$ is feature set.
Each feature contains 2D observation and its BRIEF descriptor (32 Byte). 
The feature descriptors cost the most memory, which equals a 500$\times$32 resolution image for 500 features in one keyframe. 
Therefore, it takes approximately 17kB for one keyframe.

\begin{figure*} 
    \centering
    \subfigure[MH01] { \label{fig:MH01} 
        \includegraphics[width=0.37\columnwidth]{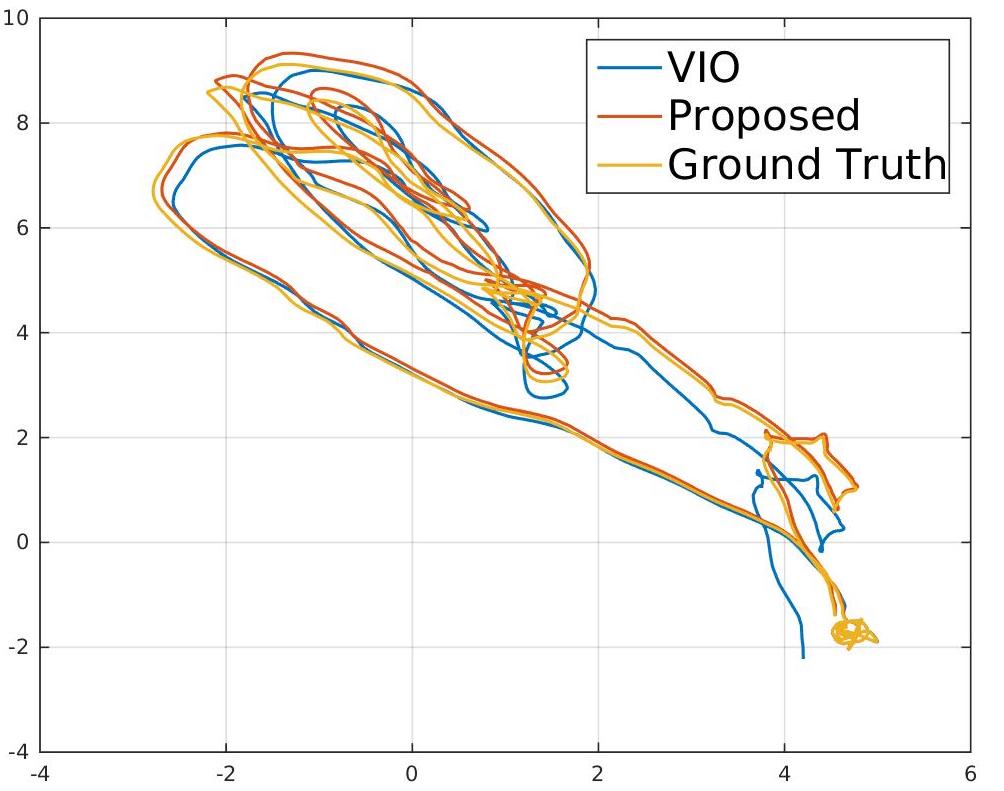}     } 
    \subfigure[MH02] { \label{fig:MH02} 
        \includegraphics[width=0.37\columnwidth]{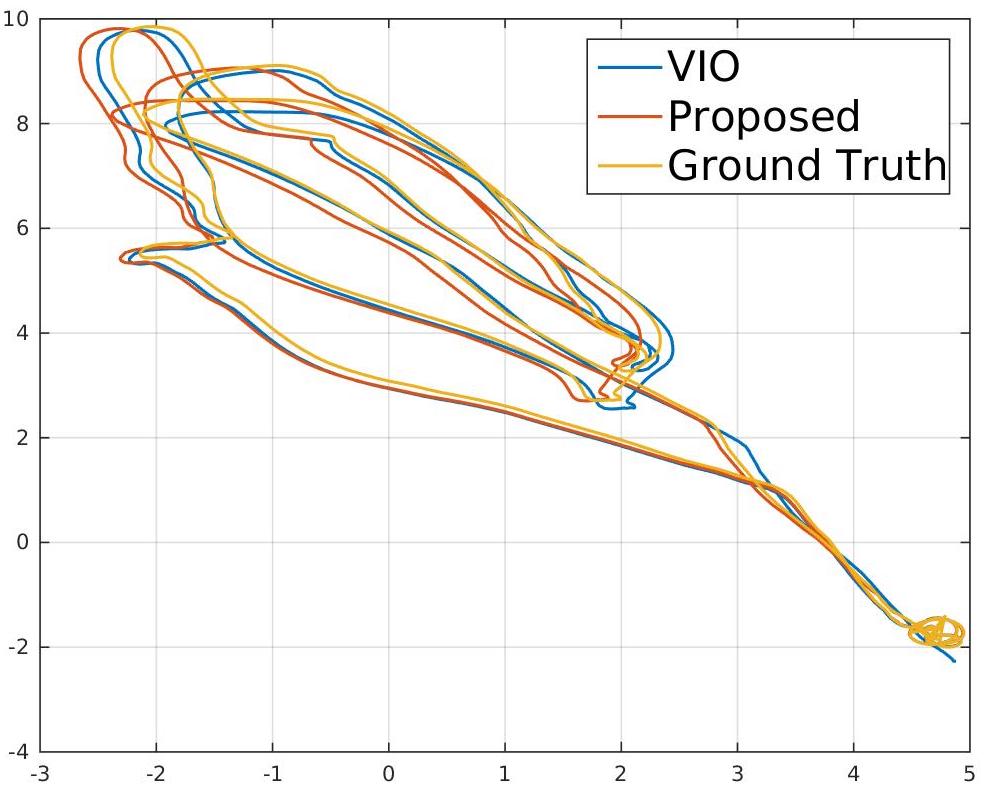} }
    \subfigure[MH03] { \label{fig:MH03} 
        \includegraphics[width=0.37\columnwidth]{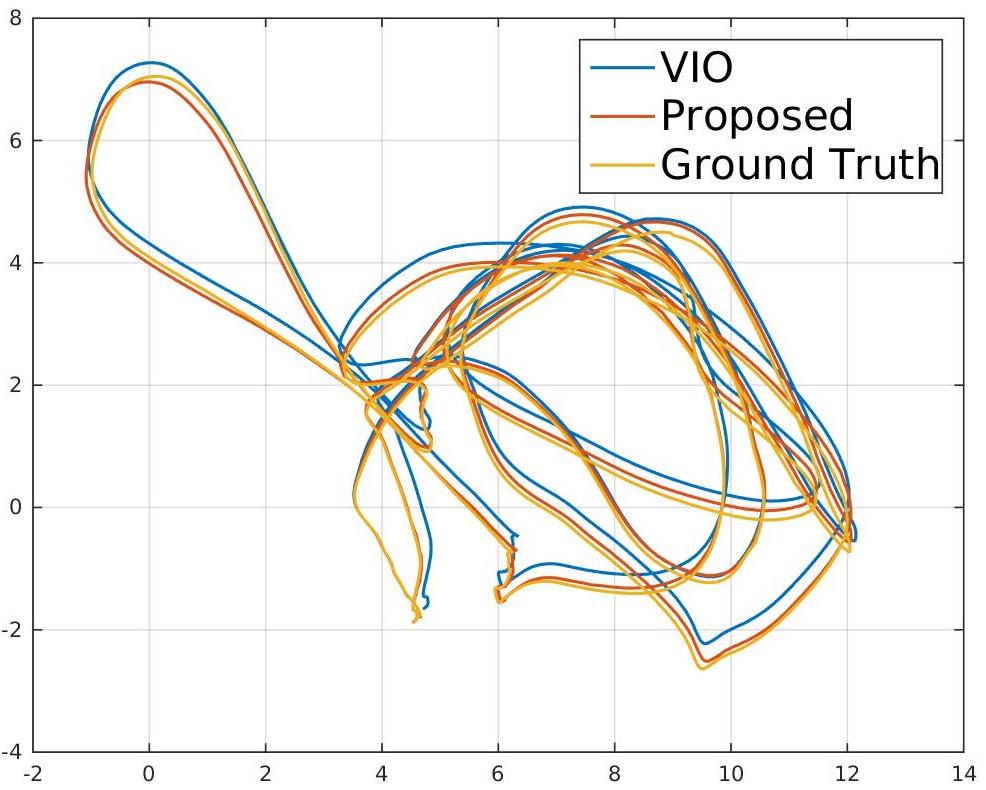} }
    \subfigure[MH04] { \label{fig:MH04} 
        \includegraphics[width=0.37\columnwidth]{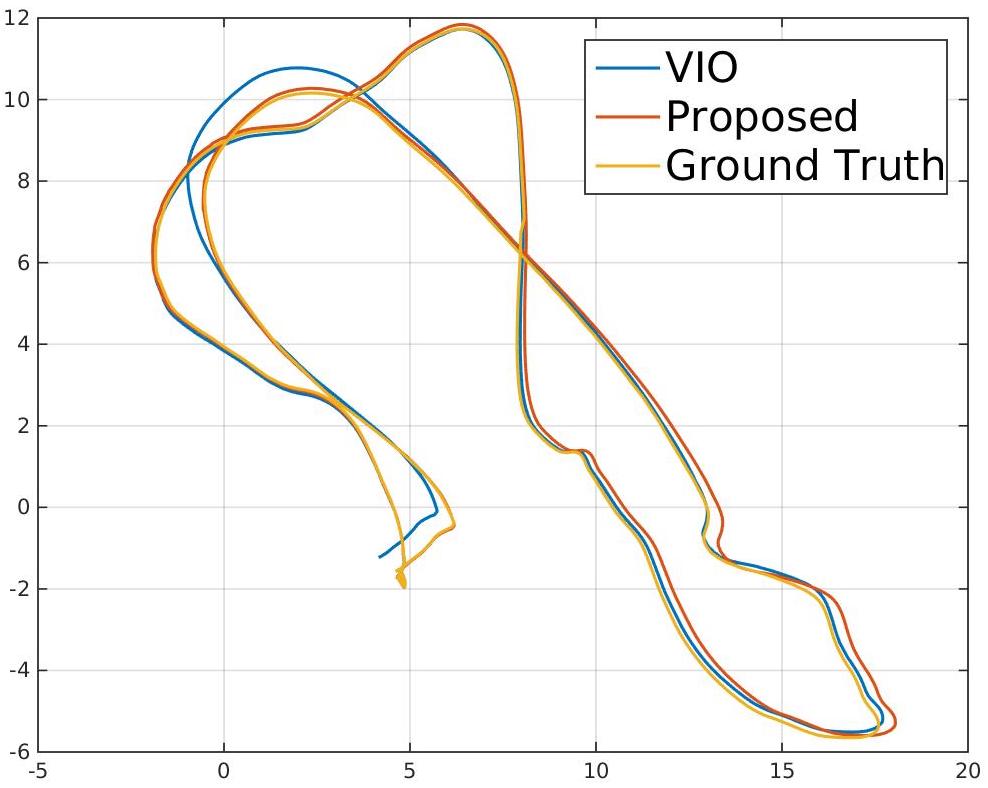} }
    \subfigure[MH05] { \label{fig:MH05} 
        \includegraphics[width=0.37\columnwidth]{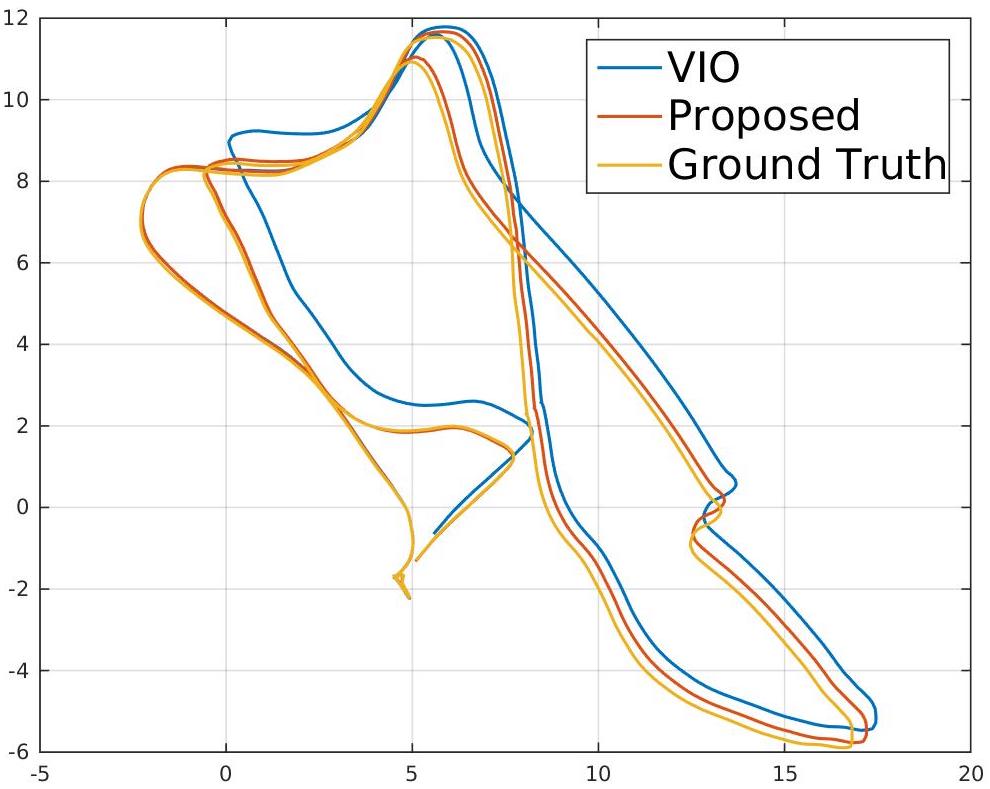} }
    \subfigure[V101] { \label{fig:V101} 
        \includegraphics[width=0.37\columnwidth]{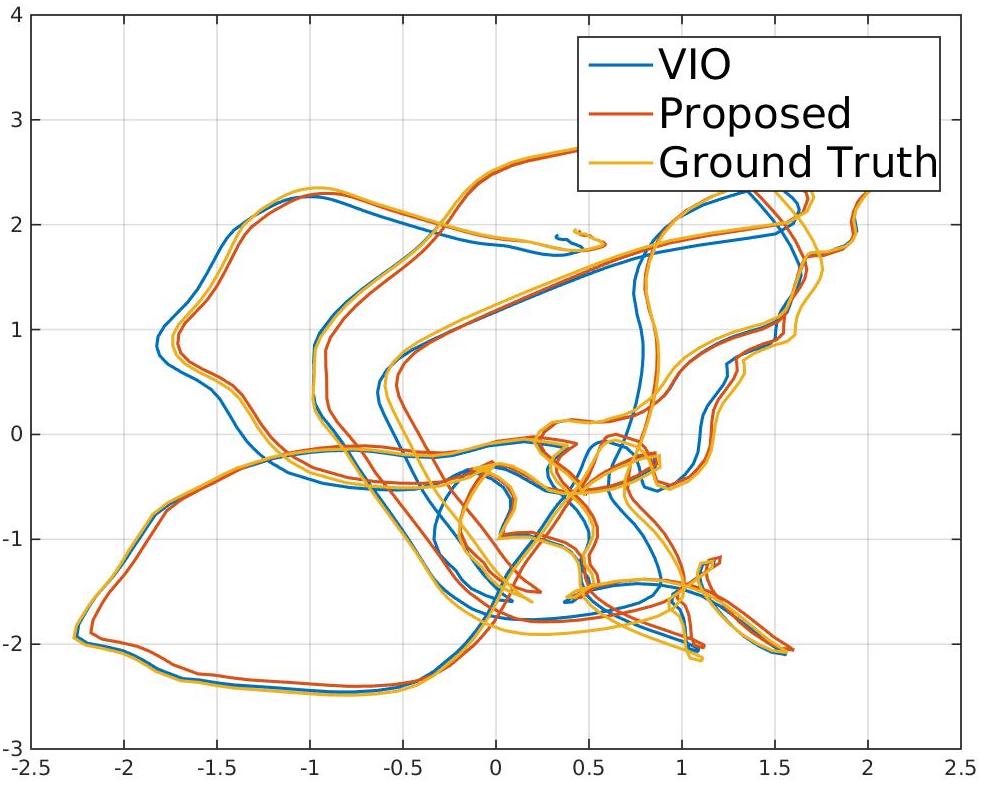} }
    \subfigure[V102] { \label{fig:V102} 
        \includegraphics[width=0.37\columnwidth]{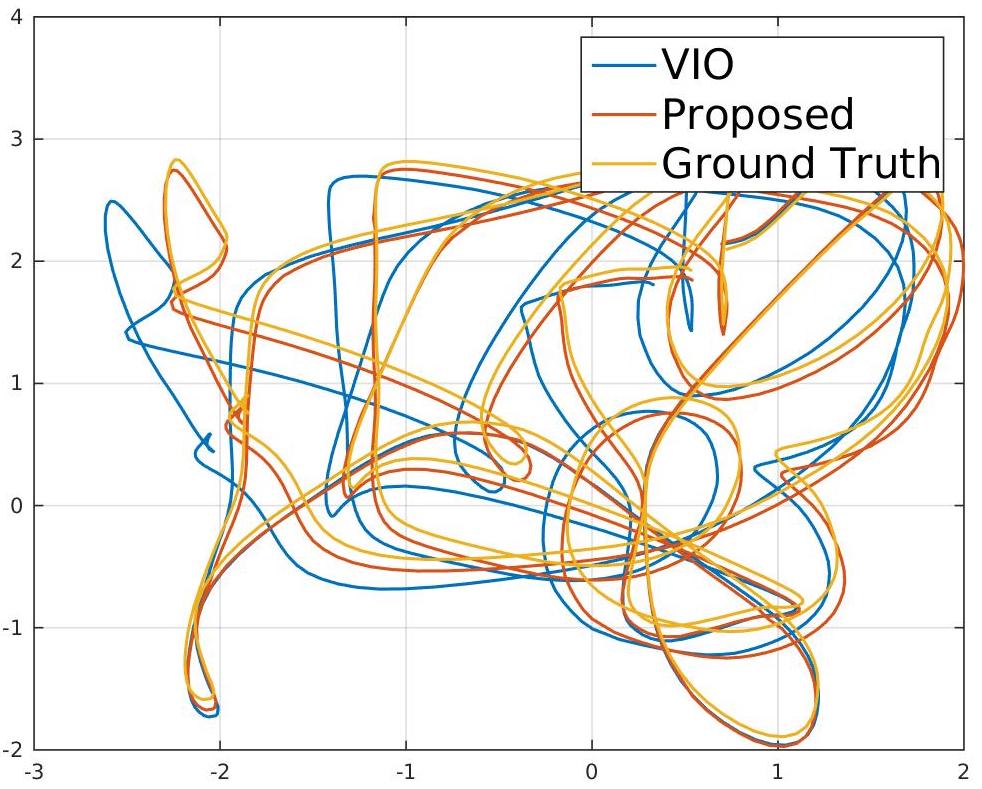}} 
    \subfigure[V103] { \label{fig:V103} 
        \includegraphics[width=0.37\columnwidth]{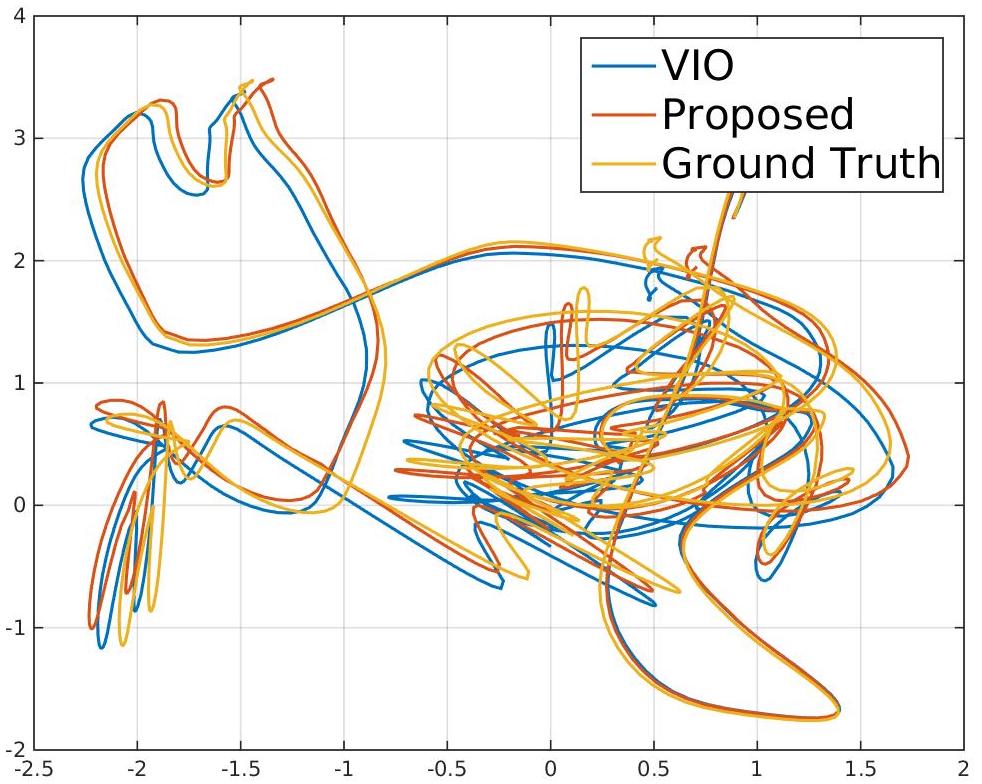} }
    \subfigure[V201] { \label{fig:V201} 
        \includegraphics[width=0.37\columnwidth]{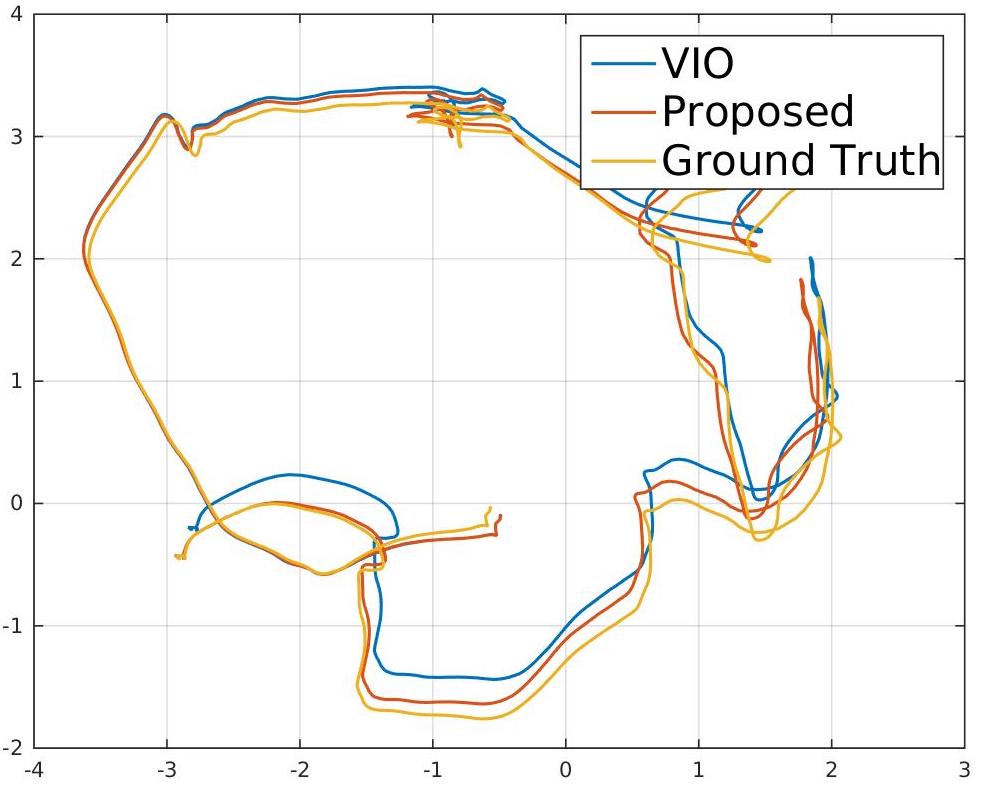} }
    \subfigure[V202] { \label{fig:V202} 
        \includegraphics[width=0.37\columnwidth]{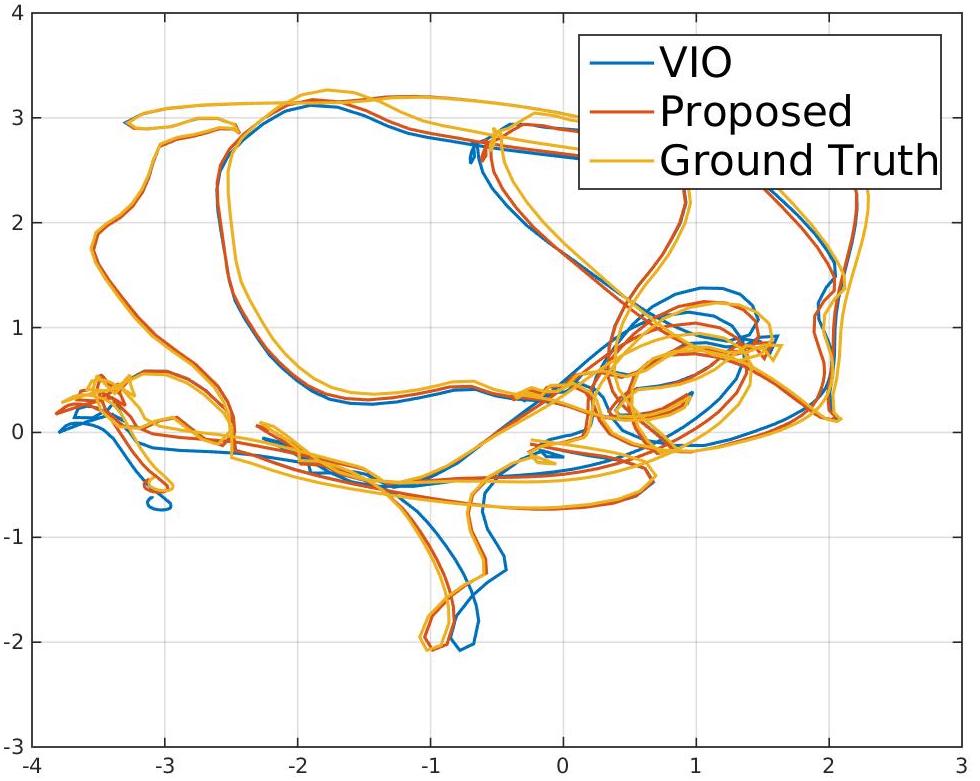} }
    
    \caption{Trajectory of proposed system in EuRoC dataset. Our system compares with ground truth and pure VIO without loop closure. } 
    \label{fig:euroc trajectory} 
\end{figure*}

\subsubsection{Pose Graph Loading}
We use the same saving format to load keyframe. 
Every keyframe is a vertex in the pose graph.
The initial pose of vertex is $\hat{\mathbf{p}}^w_i,\hat{\mathbf{p}}^w_i$. 
The loop edge is established directly by the loop information $\hat{\mathbf{p}}^i_{iv}, \hat{\psi}_{iv}$.
Every keyframe establishes several sequential edges with its neighbor keyframes, as eq. \eqref{eq:loop constraint}. 
After loading the pose graph, we perform global 4-DOF pose graph once immediately. 
The speed of pose graph saving and loading is in linear correlation with pose graph's size.

\begin{table*}
    \centering
    \caption{Absolute trajectory error, ATE \cite{sturm2012benchmark} in EuRoC datasets in meters. \label{tab:euroc_error}}
    \begin{tabular}{cccccccccccc}
        \toprule
        Sequence & MH01 & MH02 & MH03 & MH04 & MH05 & V101 & V102 &V103 &V201 &V201 & MH\_all\_merged \\
        \midrule
        VIO without loop      & 0.15  & 0.15 & 0.22 & 0.32 & 0.30 & 0.079 & 0.11 & 0.18 & 0.080 & 0.16 & \\
        VI SLAM\cite{kasyanov2017keyframe}  & 0.25  & 0.18 & 0.21 & 0.30 & 0.35 & 0.11 & 0.13 & 0.20 & 0.12 & 0.20  & \\
        proposed       & \textbf{0.12}  & \textbf{0.12} & \textbf{0.13} & \textbf{0.18} & \textbf{0.21} & \textbf{0.068} & \textbf{0.084} & 0.19 & 0.081 & \textbf{0.16} & \textbf{0.21}\\
        \bottomrule
    \end{tabular}
\end{table*}

\section{Experiment Results}
\label{sec:experiments}

We validate proposed system on a public dataset and outdoor environment. 
In the public dataset experiments, we compare the proposed algorithm with another state-of-the-art algorithm \cite{kasyanov2017keyframe}. 
A numerical analysis shows the accuracy of our proposed system. 
We also merge different sequences into a global pose graph.
The outdoor experiment is performed to illustrate the large-scale practicability of our system.

\subsection{Public Dataset}

\begin{figure}
    \centering
    \includegraphics[width=0.4\textwidth]{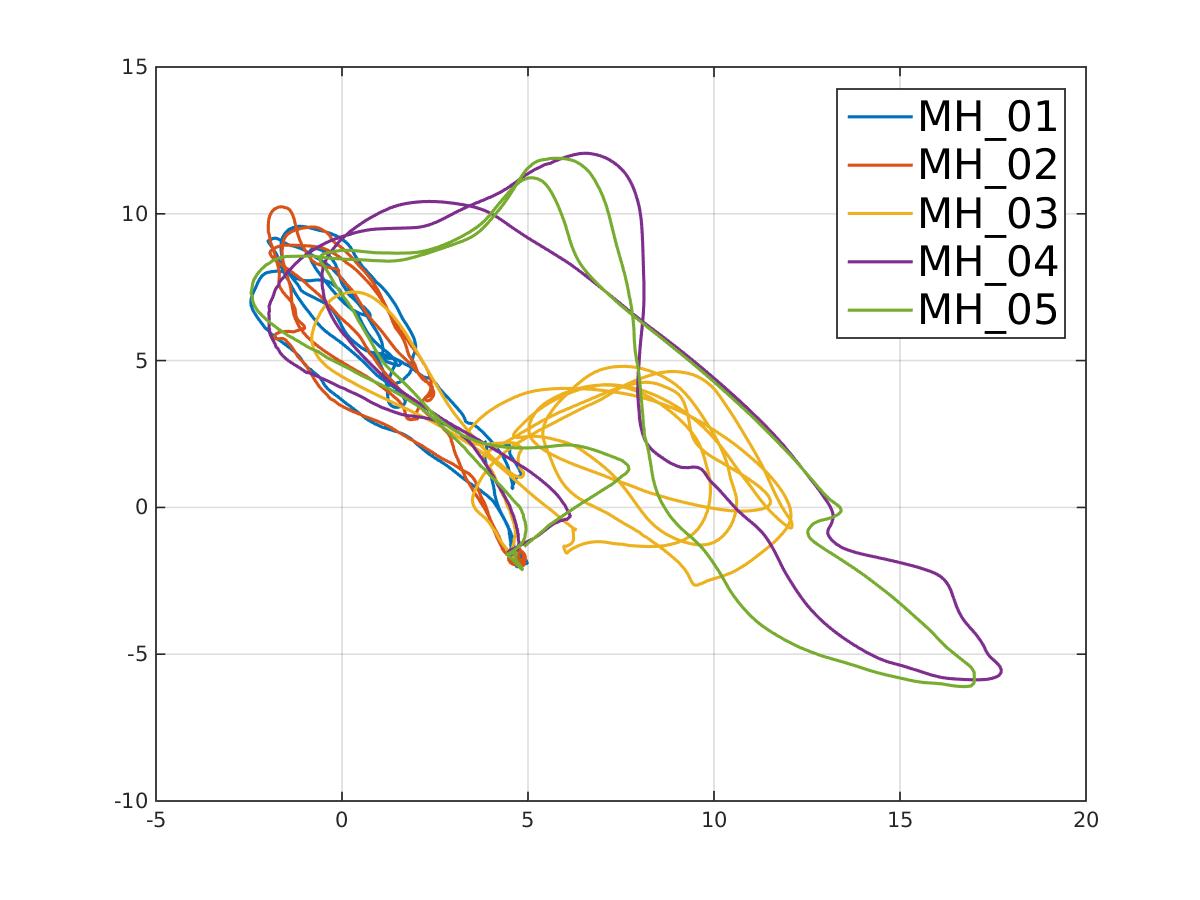}
    \caption{Trajectories of all Machine hall sequences in a global map.
        \label{fig:MH_all_trajectory} }
     \vspace{-0.8cm}
\end{figure}

\begin{figure}
    \centering
    \includegraphics[width=0.4\textwidth]{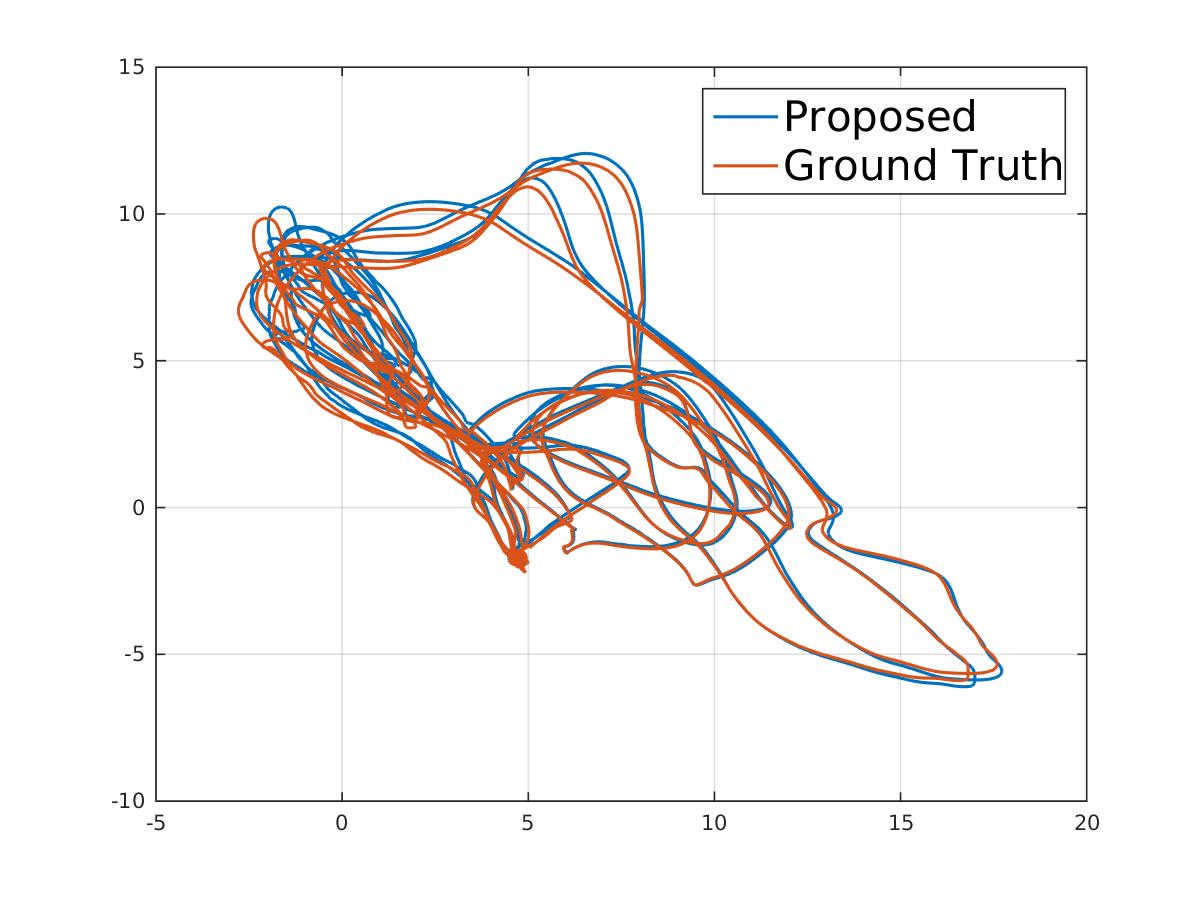}
    \caption{Trajectories merging results of our system compare against with ground truth.
        \label{fig:MH_all_compare} }
\end{figure}

We evaluate proposed system using the EuRoC MAV Visual-Inertial Datasets\cite{Burri25012016}. 
The datasets are collected onboard a micro aerial vehicle, which contains stereo images (Aptina MT9V034 global shutter, WVGA monochrome, 20 FPS), 
synchronized IMU measurements (ADIS16448, 200 Hz), and ground truth states (VICON and Leica MS50). 
The datasets contain 5 sequences in the machine hall and 6 sequences in the Vicon room.
The ground truth is provided by the laser tracker and motion capture system respectively. 
Only images from the left camera are used in experiments. 

The trajectory of all the sequences is shown in Fig. \ref{fig:euroc trajectory}.
We compare the proposed system with ground truth and VIO without loop closure \cite{QinShen17} in one figure.
It can be seen that our relocalization and global pose graph optimization greatly increase the accuracy of pure VIO.
For quantitative analysis, we compare our system against another state-of-the-art SLAM work, VI SLAM \cite{kasyanov2017keyframe}. which is built on the top of OKVIS \cite{LeuFurRab1306}. 
VI SLAM relocalize camera pose only between two frames.
It performs 6-DOF pose graph optimization after relocalization.
We compare quantitative results in terms of absolute trajectory error (ATE,\cite{sturm2012benchmark}).
As shown in Table. \ref{tab:euroc_error}, our relocalization and global pose graph optimization improve pure VIO result obviously. 
Furthermore, our proposed system outperforms VI SLAM \cite{kasyanov2017keyframe} in the most of sequences. 
Because our relocalization performs in a tightly-coupled local window instead of only two frames, our relocalization results are more accurate. 
In addition, our 4-DOF optimization seizes the actual drifted direction, ignoring drift-free directions, which corrects drift more effectively and accurately. 

We also merge five Machine hall sequences into one map.
To the best of our knowledge, this is the first work trying to splice different visual-inertial sequences together on EuRoC dataset. 
These sequences start at different poses and different times. 
We do relocalization and pose graph optimization based on similar camera views in diffident sequences. 
We only fix the first frame in the first sequence, whose position and yaw angle is set to zero. Then we merge new sequences into previous map one by one. The trajectory is shown in Fig.~\ref{fig:MH_all_trajectory}. We also compare the whole trajectory with ground truth. The ATE is 0.34m, which is an impressive result in a 500-meter-long run in total. 
This experiment shows that the map "evolves" overtime by incrementally merging new sensor data captured at different times.

\subsection{Large-scale Outdoor Environment}
\begin{figure}
    \centering
    \includegraphics[width=0.35\textwidth]{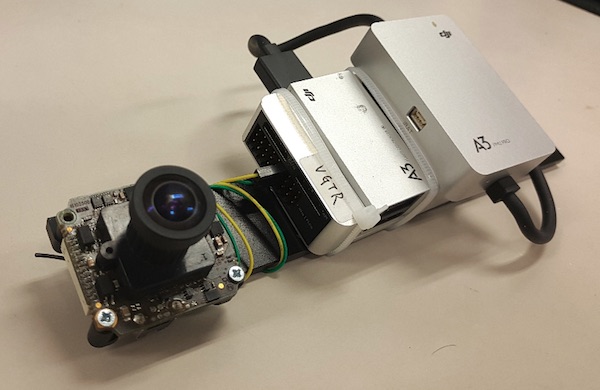}
    \caption{The device we used for the indoor experiment. 
        It contains one global shutter camera (MatrixVision mvBlueFOX-MLC200w) with 752$\times$480 resolution.
        We use the built-in IMU (ADXL278 and ADXRS290, 400Hz) from the DJI A3 flight controller. 
        \label{fig:device} }
     \vspace{-0.2cm}
\end{figure}

\begin{table}[h]
    \centering
    \caption{Timing table in outdoor experiment.\label{tab:out_time}}
    \begin{threeparttable}
        \begin{tabular}{cccc}
            \toprule
            Thread &  Modules  & Time(ms) & Rate\\
            \midrule
            1 & VIO        & 40  &10 Hz\\
            \midrule
            2 & Loop Detection  & 8 & every keyframe  \\
            2 & Relocalization  & 40 & every loop     \\
            2 & Pose Graph Optimization & 186 & every loop \\
            \midrule
            & Pose Graph Save & 907 & once\\
            & Pose Graph Load & 4464& once\\
            \bottomrule
        \end{tabular}
        \begin{tablenotes}
            \footnotesize
            \item[*] This table represents maximum time cost in the outdoor experiment, which has 2747 keyframes in the pose graph.  
        \end{tablenotes}
    \end{threeparttable}
    \vspace{-0.3cm}
\end{table}

\begin{figure}
    \centering
    \includegraphics[width=0.35\textwidth]{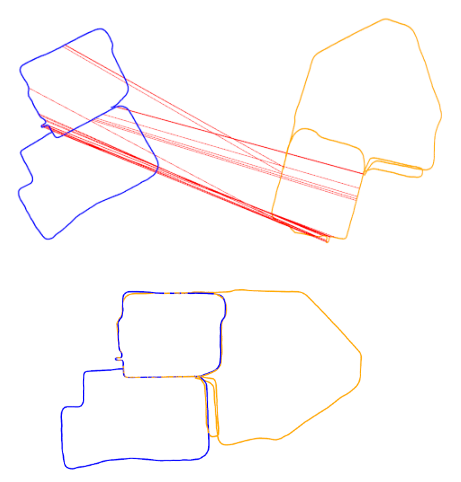}
    \caption{The left picture shows the trajectories of two sequences respectively. The yellow line is the trajectory of sequence 1 and the blue line is the trajectory of sequence 2. The red lines, connecting two trajectories, draw loop detection places. 
        The right picture shows the merging results. 
        \label{fig:outdoor_merge} }
\end{figure}

\begin{figure}
    \centering
    \includegraphics[width=0.45\textwidth]{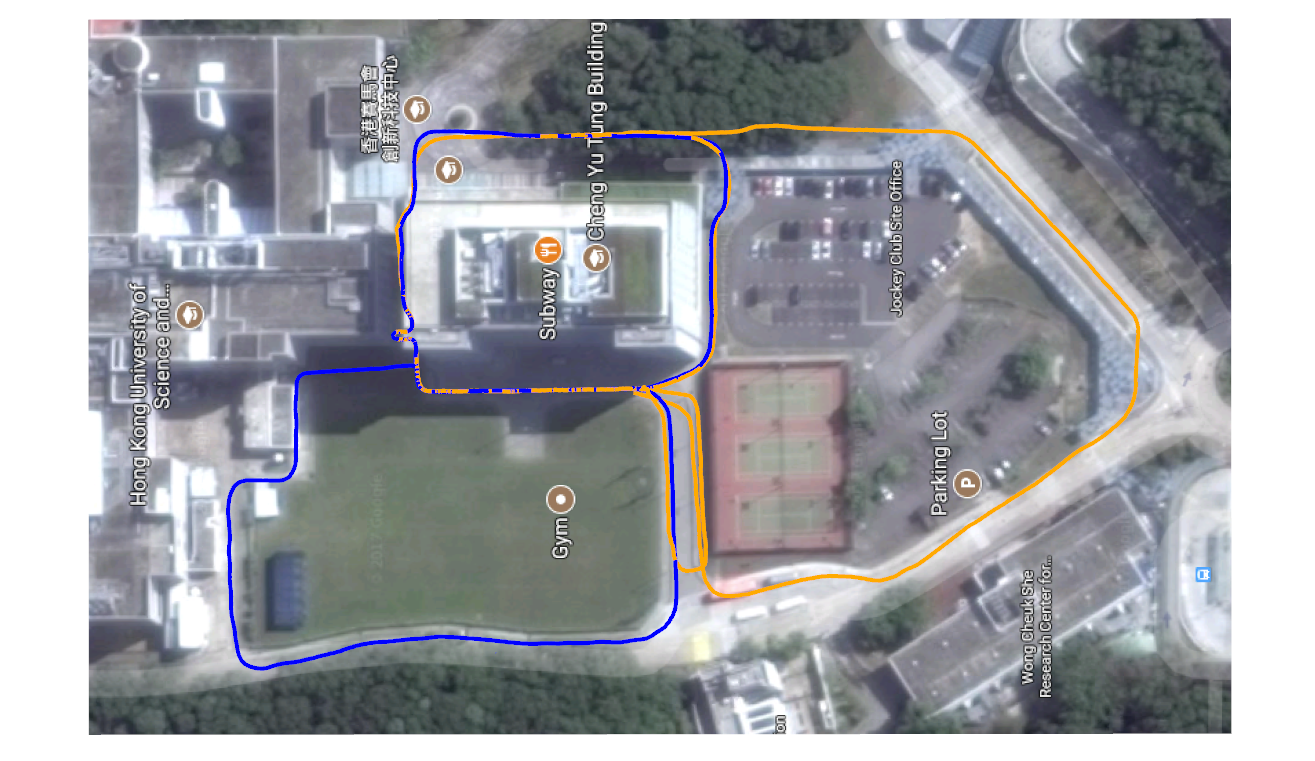}
    \caption{The whole trajectory aligns with Google Map.
        \label{fig:map_merge_result} }
\end{figure}

This experiment valid proposed relocalization and pose graph optimization in the large-scale outdoor environment.

The sensor suite used in this experiment is shown in Fig.~\ref{fig:device}.
It contains a monocular camera (MatrixVision mvBlueFOX-MLC200w, 20Hz, 752$\times$480 resolution) and an IMU (ADXL278 and ADXRS290, 400Hz) inside the DJI A3 controller\footnote{\url{http://www.dji.com/a3}}. 
The camera and IMU are hardware-synchronized. 
The intrinsic parameter of the camera is calibrated offline. 
The extrinsic parameter between camera and IMU is calibrated online.
The two outdoor sequences are collected by a person walking on the campus at different times. 
The first sequence is around 740 m, and the second sequence is 540m. 
We first build and save the map of sequence one. 
Then we load this map into memory.
The sequence two starts with an arbitrary unknown position.
Every keyframe is used to detect loop with the previous-built map.
Once a loop is detected, we do relocalization and global pose graph optimization to fuse this two map together. 

We process data on a desk computer equipped with an Intel(R) Core(TM) i7-3770 CPU @ 3.40GHz. 
The timing table is shown in the \ref{tab:out_time}. 
The whole system runs in the real time. 
We do loop detection for every new-coming keyframes.
Only when new loop is detected, we perform relocalization and pose graph optimization,

The trajectory is shown in Fig. \ref{fig:outdoor_merge}. The top picture in the Fig. \ref{fig:outdoor_merge} shows the trajectories of two sequences respectively. The yellow line is the trajectory of sequence 1, which serves as a previous-built map. The blue line is the trajectory of sequence 2, which serves as a current local map. The red lines, connecting two trajectories, represent loop connection between two maps. 
The bottom picture in the Fig. \ref{fig:outdoor_merge} shows the merging results. 
The two map tightly integrated together by global pose graph optimization.
For intuitive visualization purpose, we align the whole trajectory with Google Map in Fig. \ref{fig:map_merge_result}. 
The trajectory matches Google Map well, which validates proposed system.

\section{Conclusion}
\label{sec:conclusion}
In this paper, we propose a monocular visual-inertial SLAM system which has the capability of relocalization and pose graph optimization to achieve global consistency in real-time when loop closure happens. 
Our system can relocalize camera position in the previous-built map and merge current map with the previous map by pose graph optimization. 
The whole system saves and loads a pose graph efficiently, which has the ability to reuse previous results.  

Our system has the potential ability of building map for a huge city. In the future, we want to collect data and built local maps in multi distributed devices. Then we merge all the local maps into a huge global map together. Finally, we can relocalize and get the absolute pose in this global map wherever you look at.

\bibliography{paper.bib}

\end{document}